 \documentclass[pmlr,twocolumn,10pt]{jmlr} % W&CP article

\usepackage{booktabs}
\usepackage{siunitx}
\usepackage{caption}
\usepackage[switch]{lineno}

\theorembodyfont{\upshape}
\theoremheaderfont{\scshape}
\theorempostheader{:}
\theoremsep{\newline}

\jmlrvolume{260}
\jmlryear{2025}
\jmlrpublished{LEAVE UNSET}
\jmlrworkshop{Machine Learning for Health (ML4H) 2025} % W&CP title

\title[]{MIMIC-RD: Can LLMs differentially diagnose rare diseases in real-world clinical settings?}

\author{%
\Name{Zilal Eiz AlDin} \Email{zelalae2@illinois.edu}\\
\addr Department of Computer Science, University of Illinois Urbana-Champaign, Champaign, IL, USA
\AND
\Name{John Wu}\\
\addr Department of Computer Science, University of Illinois Urbana-Champaign, Champaign, IL, USA
\AND
\Name{Jeffrey Paul Fung}\\
\addr University of Illinois College of Medicine, Peoria, IL, USA
\AND
\Name{Jennifer King}\\
\addr University of Illinois College of Medicine, Peoria, IL, USA
\AND
\Name{Mya Watts}\\
\addr University of Illinois College of Medicine, Peoria, IL, USA
\AND
\Name{Lauren O'Neill}\\
\addr University of Illinois College of Medicine, Peoria, IL, USA
\AND
\Name{Adam Richard Cross}\\
\addr University of Illinois College of Medicine, Peoria, IL, USA
\AND
\Name{Jimeng Sun} \Email{jimeng@illinois.edu}\\
\addr Department of Computer Science, University of Illinois Urbana-Champaign, Champaign, IL, USA
}

\begin{document}

\maketitle

\begin{abstract}
Despite rare diseases affecting 1 in 10 Americans, their differential diagnosis remains challenging. Due to their impressive recall abilities, large language models (LLMs) have been recently explored for differential diagnosis. Existing approaches to evaluating LLM-based rare disease diagnosis suffer from two critical limitations: they rely on idealized clinical case studies that fail to capture real-world clinical complexity, or they use ICD codes as disease labels, which significantly undercounts rare diseases since many lack direct mappings to comprehensive rare disease databases like Orphanet.
To address these limitations, we explore MIMIC-RD, a rare disease differential diagnosis benchmark constructed by directly mapping clinical text entities to Orphanet. Our methodology involved an initial LLM-based mining process followed by validation from four medical annotators to confirm identified entities were genuine rare diseases.
We evaluated various models on our dataset of 145 patients and found that current state-of-the-art LLMs perform poorly on rare disease differential diagnosis, highlighting the substantial gap between existing capabilities and clinical needs. From our findings, we outline several future steps towards improving differential diagnosis of rare diseases.
% Rare diseases affect 1 in 10 Americans, yet standard ICD coding systems fail
% to capture these conditions in electronic health records (EHR), leaving crucial
% information buried in clinical notes. Current approaches struggle with medical
% abbreviations, miss implicit phenotype mentions, raise privacy concerns with
% cloud processing, and lack clinical reasoning abilities. We present Rare Dis-
% ease Mining Agents (RDMA), a framework that mirrors how medical experts
% identify rare disease patterns in EHR. RDMA (i) loads de-identified MIMIC-
% IV notes using PyHealth, (ii) runs on-prem Slurm jobs to extract → verify →
% match rare-disease entities, then extract → verify → match HPO phenotypes,
% (iii) aggregates multi-annotator labels to construct high-agreement gold sets, and
% (iv) benchmarks multiple LLMs via HitK differential-diagnosis evaluation from
% phenotypes. On a consensus subset, RDMA alone achieves moderate agreement
% with gold labels, and RDMA + human-in-the-loop reaches near-perfect agree-
% ment by Cohen’s K(0.81). 
\end{abstract}

\paragraph*{Data and Code Availability}
We use MIMIC-IV \citep{johnson2023mimic4, johnson2023mimic4_note} for our rare disease benchmark construction. We offer our initial mining scripts through \url{https://github.com/zelal-Eizaldeen/rare_disease_pyHealth/tree/main} and benchmark data \url{https://github.com/jhnwu3/RDMA/blob/main/public_data/initial_diff_diagnosis_benchmark.json}.

\paragraph*{Institutional Review Board (IRB)} 
We received IRB approval from the relevant institutions to allow medical students to annotate whether a text entity was a rare disease or not.
% needs to be double blind unfortunately
% Code Available at \cite{jzcode25}

\section{Introduction}
\label{sec: intro}
\begin{figure*}[h]
    \centering
    \vspace{-0.2cm}
    \includegraphics[width=1.0\textwidth]{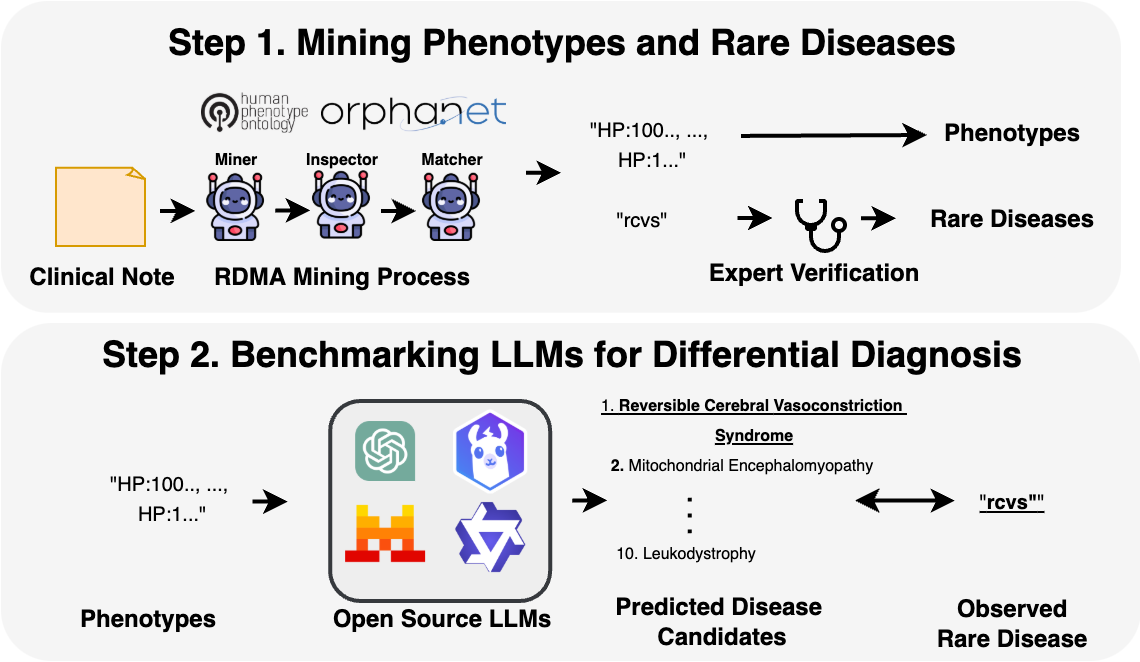}
    \caption{We directly mine rare disease mentions from clinical notes along with any phenotypes using RDMA \citep{wu2025rdmacosteffectiveagentdriven} and we verify its mined outputs directly across 4 medical students, subsampling rare disease mentions with more than 3 annotators in agreement. We use this mining to benchmark a variety of LLM models for differential diagnosis. of rare diseases}
    \label{fig:MiningToDiffDiagnosis}
    \vspace{-0.2cm}
\end{figure*}

Rare diseases affect approximately 1 in 10 Americans, constituting a significant healthcare challenge despite their individual rarity \citep{VTNews2025_general_fact}. Accurate diagnosis remains difficult due to the vast diversity and sparsity of these conditions \citep{auvin2018problem_rarity}. Large language models (LLMs) are being actively explored for differential diagnosis due to their wide-recall abilities \citep{mcduff2025towards}, leading to a plethora of work on LLM-based rare disease differential diagnosis. These approaches range from exploring large frontier models \citep{ao2025comparative_LLM_rd, chen2024rarebenchllmsserverare} to smaller agentic systems \citep{chen2025rareagentsadvancingraredisease}.

However, existing approaches suffer from several drawbacks. Studies by \citet{ao2025comparative_LLM_rd} and \citet{chen2024rarebenchllmsserverare} focus purely on cleanly-curated clinical case studies with patient profiles that can differ vastly from typical clinical settings, such as those found in clinical notes from MIMIC-IV \citep{johnson2023mimic4, johnson2023mimic4_note}. Meanwhile, \citet{chen2025rareagentsadvancingraredisease} use ICD codes as a proxy for rare disease classification. While efforts have been made to map ICD codes to more granular rare disease Orphanet codes \citep{cavero2020icd10_to_orpha}, over 50\% of Orphanet codes lack direct mappings, resulting in under-reporting of rare diseases within ICD-annotated systems.

To ensure better coverage, it is crucial to directly mine rare diseases from clinical notes and map them to specialized rare disease ontologies like Orphanet \citep{weinreich2008orphanet, mazzucato2023orphacodes_better}. Fortunately, recent advancements in phenotype extraction with LLMs \citep{garcia2024_HPORAG} and rare disease mention extraction, such as RDMA \citep{wu2025rdmacosteffectiveagentdriven}, have achieved high precision results. Leveraging RDMA, we mine phenotypes mapped to the Human Phenotype Ontology (HPO) and rare diseases mapped to the Orphanet ontology. We validate the rare disease mentions (explicitly defined by Orphanet) with four medical students to ensure only true positive cases exist in our benchmark.

Our findings demonstrate that:
\begin{itemize}
    \item LLMs severely underperform in rare disease differential diagnosis despite higher reported performance in prior work \citep{chen2025rareagentsadvancingraredisease}
    \item Key challenges exist in the relationship between patient phenotype presentation and the ability to correctly predict rare diseases
\end{itemize}

We hope this work serves as an entry point for better discussions surrounding the use of LLMs for differential diagnosis as a key component of future healthcare research.
\section{Methodology}
\textbf{Mining rare diseases and phenotypes.} We use RDMA \citep{wu2025rdmacosteffectiveagentdriven} to extract rare disease mentions from clinical notes. The approach follows a three-step process: LLMs first extract potential rare disease entities, then verify whether they represent actual rare diseases, and finally map them to their respective ontologies. To balance reliability and diversity in our entity capture, we run this pipeline twice—once at low temperature ($T=0.01$) for consistent results and once at high temperature ($T=0.7$) for broader coverage. 

Next, four medical students manually verify each extracted rare disease mention within its clinical context, determining whether it truly represents a rare disease according to Orphanet criteria \citep{weinreich2008orphanet}. Annotators make binary "yes or no" decisions on whether each mined entity constitutes a genuine rare disease observation for the patient. We then create a high-agreement subset for final evaluation, requiring consensus from at least three annotators.

For patients with verified rare disease mentions, we extract phenotypes using the same three-step RDMA pipeline, mapping them directly to the Human Phenotype Ontology \citep{gargano2024mode_hpo}. As a final preprocessing step, we address cases where entities map to both Orphanet and the Human Phenotype Ontology, making them both rare diseases and phenotypes. We remove these overlapping phenotypes (less than 2.5\% of cases) from each patient's phenotype set to avoid redundancy. 

\textbf{Differential diagnosis setup.} Effectively, given a patient's list of observed phenotypes, LLMs are asked to predict a list of the 10 most likely rare diseases in order of likelihood. To benchmark their performance, we check if an observed rare disease falls within their top $k$ predictions, hence the Hit@k metric ($k=1,5,10$).

\section{Results}
\textbf{Baselines.} We explore 5 readily available open source models. OpenBio-LLM 70B \citep{OpenBioLLMs}, Qwen 3 32B \citep{yang2025qwen3}, Llama 3.3 70B \citep{grattafiori2024llama3}, Mistral 3.1 24B \citep{mistral2025small}, and Mixtral 70B \citep{jiang2024mixtral}.

\begin{figure}[h]
    \centering
    \vspace{-0.2cm}
    \includegraphics[width=0.5\textwidth]{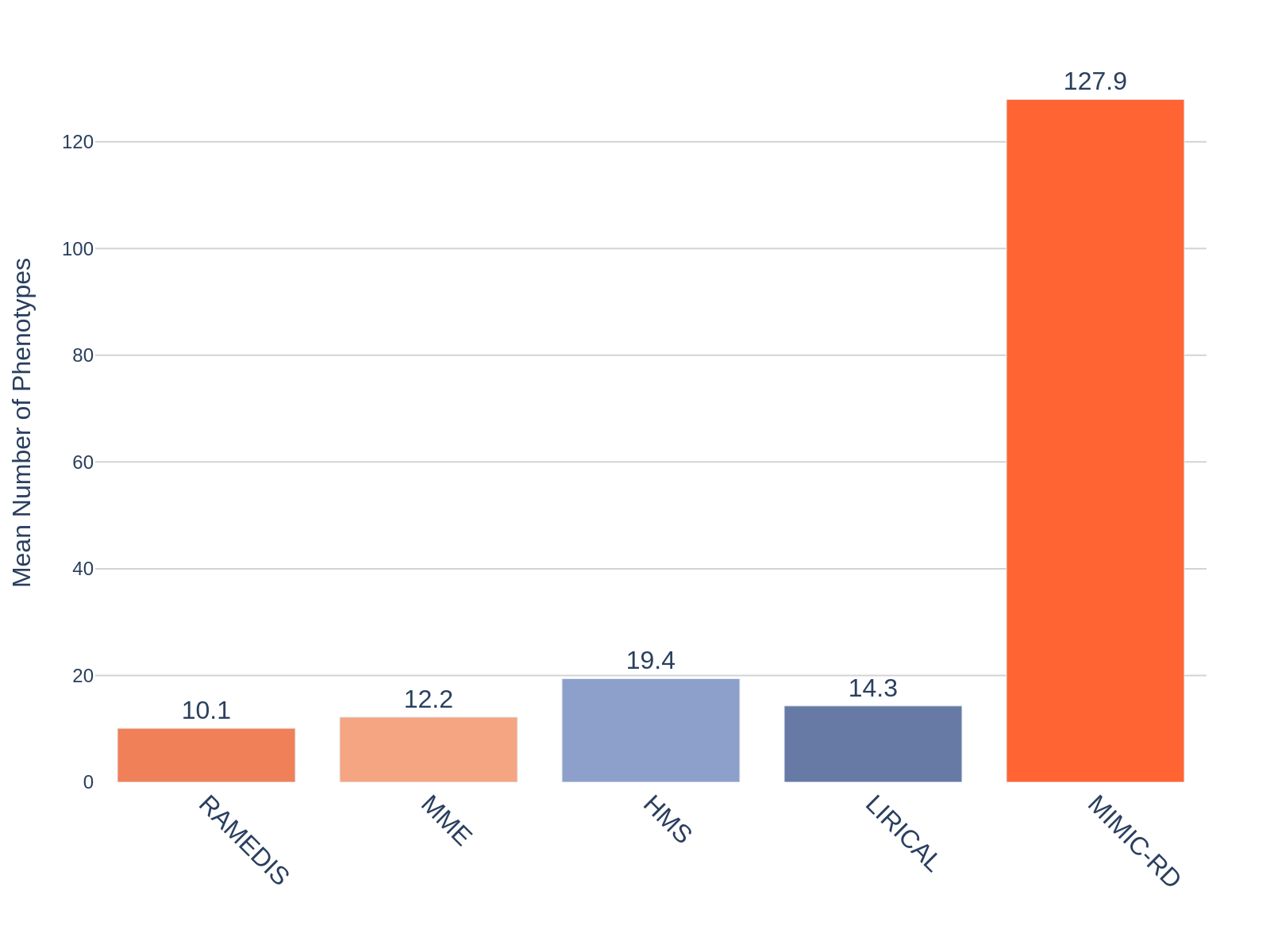}
    \caption{Mined from clinical notes with thousands of words, MIMIC-RD offers substantially greater numbers of phenotypes per patient for each rare disease. As a key implication, this dramatic increase in observations makes differential diagnosis a more complicated task as LLMs have to consider upwards of 10x more phenotypes in its differential diagnosis.}
    \label{fig:PhenotypeDatasetComparison}
    \vspace{-0.2cm}
\end{figure}

\textbf{Real-world clinical setups are substantially more complex.} We sampled 1,000 patients from MIMIC-IV \citep{johnson2023mimic4} and mined their clinical notes for rare disease mentions. Using RDMA, we identified 223 patients who contained rare disease references. Our annotators then reviewed these cases and confirmed rare diseases in 145 patients with high inter-rater agreement.
From the discharge summaries of these 145 confirmed cases, we extracted associated phenotypes. Our benchmark contains an average of approximately 128 phenotypes per patient—substantially more than existing public benchmarks including RAMEDIS \citep{topel2010ramedis}, MME \citep{philippakis2015matchmaker_mme}, HMS \citep{ronicke2019can_hms}, and Lirical \citep{robinson2020interpretable_lirical}, as shown in Figure \ref{fig:PhenotypeDatasetComparison}. In contrast with our mined phenotypes, our dataset contains far fewer observed rare diseases, averaging approximately 1 per patient.

\begin{table}[h!]
\centering
\caption{MIMIC-RD Statistics: We observe high inter-annotator agreement, a massive phenotype to disease ratio in the MIMIC hospital setting. }
\begin{tabular}{lc}
\toprule
\textbf{Statistic} & \textbf{Value} \\
\midrule
\# of Patients & 145 \\
\# of Unique Diseases & 120 \\
\# of Diseases & 192 \\
Avg. \# of Phenotypes Per Patient & 127.93 \\
Avg. \# of Rare Diseases Per Patient & 1.32 \\
Mean Pairwise Cohen Kappa & \textbf{0.71} \\ 
Fleiss' Kappa & \textbf{0.71} \\
\bottomrule
\end{tabular}
\end{table}

\textbf{Open-source LLMs still highly underperform in differential diagnosis rare disease tasks.} Even when phenotypes and rare diseases are extracted from the same clinical note, state-of-the-art open-source LLMs fail to predict approximately 60\% of observed rare diseases given their relevant phenotypes. Our rare disease hit-rates in Tables \ref{tab:DisLevelRDD} and \ref{tab:PatLevelRDD} are substantially lower than those reported in \citep{chen2025rareagentsadvancingraredisease}, showing approximately 35\% lower Hit@10 scores.

\begin{table}[h!]
\centering

\begin{tabular}{lccc}
\toprule
Model & Hit@1 & Hit@5 & Hit@10 \\
\midrule
Llama 3.3 70B          & \textbf{20.31\%} & \textbf{35.94\%} & \textbf{40.10\%} \\
Mistral 24B         & 13.02\% & 27.60\% & 33.33\% \\
OpenBioLLM 70B      & 9.47\% & 17.89\% & 26.32\% \\
Mixtral 70B         & 6.84\% & 15.79\% & 22.63\% \\
Qwen 32B  & 18.23 \% & 30.20 \% & 37.50 \% \\
\bottomrule

\end{tabular}
\caption{Differential-diagnosis recall performance (Hit@k) across all 192 observed diseases. Llama 3.3 \citep{grattafiori2024llama3} emerges as a remarkably strong performer in rare disease differential diagnosis, substantially outperforming its biomedically fine-tuned Llama 3 counterpart \citep{OpenBioLLMs}. This suggests that biomedical fine-tuning does not necessarily improve performance on related tasks such as rare disease differential diagnosis. }
 \label{tab:DisLevelRDD}
\end{table}

% Patient-level metrics table
\begin{table}[h!]
\centering

\begin{tabular}{lccc}
\toprule
Model & Hit@1 & Hit@5 & Hit@10 \\
\midrule
Llama 3.3 70B          & \textbf{19.31\%} & \textbf{30.34\%} & \textbf{35.17\%} \\
Mistral 24B         & 13.79\% & 24.83\% & 31.03\% \\
OpenBioLLM 70B      & 11.03\% & 17.93\% & 25.52\% \\
Mixtral 70B         & 7.59\% & 15.17\% & 20.69\% \\
Qwen 32B         & 18.62\% & 27.59\% & 34.48\% \\
\bottomrule

\end{tabular}
\caption{Differential-diagnosis recall performance (Hit@k) across 145 patients. Llama 3.3 \citep{grattafiori2024llama3} again emerges as a remarkably strong performer in rare disease differential diagnosis, substantially outperforming its biomedically fine-tuned Llama 3 counterpart \citep{OpenBioLLMs}. At the patient level, the relative performance ranking across models remains consistent.}      
 \label{tab:PatLevelRDD}
\end{table}

\textbf{Patient profiles typically have at least one overlapping phenotype with a rare disease profile.} Regardless of whether an LLM correctly proposes a rare disease within its top 10 candidates, patient phenotype profiles typically contain at least one phenotype that appears in the corresponding rare disease's profile on Orphanet \citep{weinreich2008orphanet}. This suggests that the majority of failed rare disease predictions are not due to missing relevant phenotype information, but rather the models' inability to appropriately rank the correct rare disease above other conditions in their top 10 predictions.

\begin{figure}[h]
    \centering
    \vspace{-0.2cm}
    \includegraphics[width=0.5\textwidth]{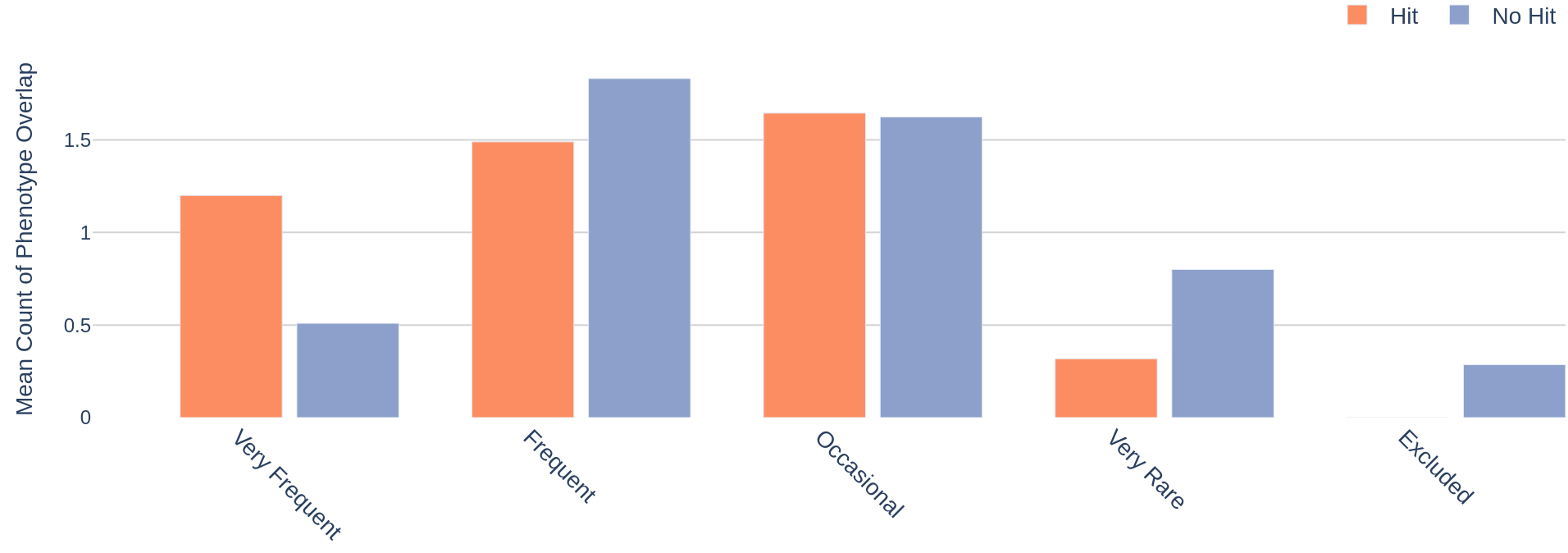}
    \caption{Comparison of phenotype overlap between "Hit@10" and "No Hit" patients with documented clinical phenotype presentations of rare diseases. Here, we plot the average number of phenotype overlaps that a patient's phenotype profile has with well-documented phenotypes for each rare disease. Such documented phenotypes have classified incidence rates to indicate their frequency of co-occurrence with a rare disease as defined by Orphanet \citep{weinreich2008orphanet} Specifically, these frequencies are rated as "very frequent" with 99-80\% co-occurrence, frequent with "79-30\%", "occasional" with "29-5\%", very rare.}
    \label{fig:PhenotypeBreakdown}
    \vspace{-0.2cm}
\end{figure}

\textbf{LLM diagnosis performance appears to be aligned with documented phenotype presentations.}  When comparing patient phenotype profiles to documented clinical presentations, we observe distinct patterns in Figure \ref{fig:PhenotypeBreakdown}: patients with correctly predicted rare diseases show substantial phenotype overlap with their disease's documented profile. In contrast, LLMs typically fail to identify rare diseases whose phenotype profiles are more ambiguous or contain phenotypes that co-occur less frequently with the observed conditions.

\section{Discussion}\label{sec: discussion}
\textbf{Mining a substantially larger phenotype-rare disease patient dataset.} LLMs enable the extraction of substantially more phenotype and rare disease information from clinical text across electronic healthcare systems. This approach can create more comprehensive benchmarks and, more importantly, improve our understanding of co-occurrence relationships between observed rare diseases and phenotypes. Additionally, building deep predictive models for differential diagnosis could help uncover nonlinear relationships between specific phenotype types and rare diseases.

\textbf{Narrowing the number of disease candidates through multimodality.} Our rare disease phenotype characterization mapping from Orphanet \citep{weinreich2008orphanet} reveals a wide range of rare diseases associated with each phenotype, from 1 to over 1,026 diseases. Given this variability, incorporating additional modalities is crucial for effective differential diagnosis to narrow down potential rare diseases. Lab tests and imaging such as X-rays can significantly reduce the number of candidate diseases. Fortunately, MIMIC-IV \citep{johnson2023mimic4} contains a wealth of different modalities within its datasets. Exploring these modalities could substantially improve diagnostic performance and assist clinicians who may lack the time to review complete patient histories.

\textbf{Differential diagnosis agents.} While initial attempts have been made to construct rare disease diagnostic agents \citep{chen2025rareagentsadvancingraredisease}, significant opportunities remain for evaluating automated differential diagnosis frameworks that can support physicians in providing more comprehensive patient care \citep{mcduff2025towards_diff_diag}.

\clearpage

\bibliography{jmlr-sample}

\appendix

% \section{First Appendix}\label{apd:first}

% This is the first appendix.

% \section{Second Appendix}\label{apd:second}

% This is the second appendix.

\end{document}